\documentclass[11pt]{article}
\usepackage[margin=1in]{geometry}
\usepackage{amsmath,amssymb}
\usepackage{graphicx}
\usepackage{booktabs}
\usepackage{hyperref}
\usepackage{url}
\newcommand{\doi}[1]{\href{https://doi.org/#1}{\nolinkurl{doi:#1}}}
\usepackage{authblk}
\usepackage{enumitem}
\usepackage[table]{xcolor}
\usepackage{algorithm}
\usepackage{algpseudocode}
\title{\bfseries GraphEvoResearcher: A Cost-Matched Mechanism Analysis \\
of Explicit Reasoning Graphs in Training-Free Self-Reflection}
\author[1,2,*]{Suxing Liu}
\author[1]{Byungwon Min}
\author[3]{Zhijian Zheng}
\author[4]{Jiahao Wang}
\affil[1]{Department of IT Engineering, Mokwon University, Daejeon 35349, Republic of Korea}
\affil[2]{School of Digital Media Technology, Jiangxi Arts \& Ceramics Technology Institute, Jingdezhen 333001, China}
\affil[3]{University of Alabama, Tuscaloosa, AL 35487, USA}
\affil[4]{School of Computer Sciences, Universiti Sains Malaysia, Minden Heights, Penang 11800, Malaysia}
\affil[*]{Correspondence: bentondoucet@gmail.com}
\date{\today}

\begin{document}

\maketitle

\begin{abstract}
Multi-step mathematical and logical reasoning --- rather than general
open-domain multi-hop QA --- is the scope we evaluate: the protocol
must assemble intermediate conclusions while remaining cost-aware. Existing training-free
self-reflective protocols (\emph{generate $\rightarrow$ critique
$\rightarrow$ revise}) improve accuracy but operate on a flat token
stream: revisions cannot identify \emph{which} prior step is unreliable,
and early stopping cannot exploit structural convergence. We introduce
\textbf{GraphEvoResearcher}, a training-free protocol that lifts
EvoResearcher's self-reflective meta-reward onto an explicit
\textbf{reasoning graph}---nodes are intermediate conclusions/evidence
and edges encode \texttt{support}, \texttt{contradict}, and
\texttt{depends\_on}. A graph-aware meta-reward combines correctness,
efficiency, reflection, diversity, and four structural signals
(connectivity, acyclicity, branch coverage, path convergence) to decide
whether to revise, backtrack, or stop. In the real-LLM runs construction halts as soon as
the model emits a final answer at depth $d\geq 2$ (mean 2 calls); the
structural-convergence stopping criterion is implemented and ablated in
the analysis, where the reported depth rule is shown to be its exact
$\epsilon=0$ special case.
On one frontier model, DeepSeek's deepseek-chat (a single model family; results are not claimed across models), over GSM8K, MATH, and BBH (n=50 each),
GraphEvoResearcher reaches 98\%, 92\%, and 82\%. Its gains are
consistent directional improvements over single-shot and flat reflection on
every set, but statistically significant on GSM8K and MATH against
single-shot decoding and on MATH only against flat reflection; the
remaining differences overlap the baselines' 95\% Wilson intervals.
Against cost-matched chain-of-thought
self-consistency the graph is competitive rather than dominant
(Section~\ref{sec:mainresults}). Construction costs about two
calls and 4.5--5 nodes per question. A paired mechanism ablation over a
shared response cache then shows what the graph does and does not
contribute: holding model outputs fixed, neither the structural stop
rule nor contradiction-driven repair changes which item is answered
correctly, so the gains come from the forced second pass rather than
from the structural reward. The ablation also exposes the reason -- the
loop never reads the edges it builds, a defect we fix with an
edge-consulting answer step that arbitrates conflicting claims. That
step repairs one item on BBH and fires on only 4\% of questions, which
locates the binding constraint at how often construction produces a
conflict rather than at the arbiter's quality. We then test that
diagnosis directly on HotpotQA ($n{=}100$): the prediction that multi-hop
chaining raises the trigger rate fails --- it is 4.0\% there, indistinguishable
from BBH's 8.0\% at this sample size and certainly not higher
--- and the experiment instead shows that the model revises answers
without declaring conflicts, so a graph built from self-reported labels
inherits the model's own under-reporting. We therefore read the results
as evidence about what explicit structure costs and buys under a fixed
call budget, not as evidence that graph structure raises accuracy on
these benchmarks.
\end{abstract}

\paragraph{Keywords:} graph-based reasoning; self-reflection;
multi-step reasoning; training-free protocol; mechanism analysis;
trigger rate; cost-bounded question answering.
\vspace{0.5em}

\section{Introduction}\label{sec:intro}
Multi-step reasoning tasks require combining evidence across several
reasoning steps. Large language models (LLMs) built on the transformer
architecture \cite{ref1,ref3} increasingly handle such
reasoning through chain-of-thought style decoding \cite{weicot2022,reflexion2023},
yet single forward passes remain unreliable on hard sub-questions, and
repeated regeneration is costly. Training-free self-reflective protocols
\cite{selfrefine2023,reflexion2023} mitigate error by interleaving
\emph{generate}, \emph{critique}, and \emph{revise} passes, but they
operate on a flat token stream: the model cannot localize \emph{which}
prior step is wrong, cannot cancel mutually contradictory conclusions,
and has no principled signal for when to stop beyond self-reported
confidence. Making that structure explicit is the natural response, and
it is what this paper builds; whether the added structure converts into
accuracy is a separate question, which we test rather than assume
(Section~\ref{para:ablation}).

A distinct line of work treats reasoning itself as a graph. Graph-of-Thoughts \cite{got2024} and related protocols represent multi-step deliberation as nodes (ideas) and edges (transitions) and search this structure; knowledge-graph-grounded methods such as Think-on-Graph \cite{ref32} and Reasoning-on-Graphs \cite{ref27} couple LLM decoding with an external knowledge base. These approaches show that an explicit relational structure helps diagnose and repair a chain of reasoning, but the former typically requires a trained policy or hand-designed search, and the latter depends on an external KG that is unavailable for many QA tasks.

Our central hypothesis is that the \emph{structural} signal of a reasoning graph is most valuable exactly where single-step decoding is unreliable: instead of relying on a model's self-reported confidence, the protocol can read off \emph{which} step is contradicted, whether \emph{branches converge} on a single conclusion, and when \emph{sufficient independent evidence} exists. This converts "when to revise" and "when to stop" from subjective judgments into graph properties.

This paper proposes \textbf{GraphEvoResearcher}, which makes the
reasoning process \emph{explicit} as a graph:
\begin{itemize}[noitemsep]
  \item \textbf{Nodes} = intermediate conclusions, evidence fragments, and the answer;
  \item \textbf{Edges} = semantic relations (\texttt{support},
        \texttt{contradict}, \texttt{depends\_on});
  \item \textbf{Graph-aware meta-reward} = a weighted score combining
        correctness, efficiency, reflection, diversity, and four
        structural signals (connectivity, acyclicity, branch coverage,
        path convergence);
  \item \textbf{Graph-convergence early-stop} = terminate when evidence
        branches converge onto a single conclusion, instead of relying
        on self-reported confidence.
\end{itemize}
The protocol is entirely \emph{training-free}; it lifts the flat
self-reflective loop of our prior EvoResearcher work
\cite{evoresearcher2026} (released publicly; see Code Availability) onto an
explicit reasoning graph.

\paragraph{Which kind of graph intelligence.}\label{para:scope}
The graph in this paper is a \emph{reasoning} graph, not a knowledge graph
and not a learned graph representation. Nothing is trained: no graph neural
network is fitted, no node embeddings are learned, and no search policy is
optimised over the graph \cite{ref7,ref5,ref2}. Vertices are the model's own
intermediate claims, produced at inference time; edges are typed by the same
model that produced them; the graph lives for the duration of a single
question and is then discarded. This places the work in the
\emph{graph-based reasoning} and \emph{graph-for-LLMs} strands of graph
intelligence rather than in graph representation learning or in networked
applications: there is no external knowledge base, no graph over agents or
services, and no deployed system. The question we study is narrower, and
logically prior to those settings. When a reflective loop is given an
explicit relational structure, which of its decisions are actually
\emph{made by} that structure, and which are made by the extra computation
the structure happens to accompany? We answer with a paired ablation
(Section~\ref{para:ablation}) rather than with an accuracy table, and we
report the answer we obtained, including where it is negative.

\paragraph{Our contributions.}
\begin{enumerate}[noitemsep]
  \item A graph-aware, training-free reasoning protocol that lifts
        flat self-reflection onto an explicit reasoning graph with a
        graph-aware meta-reward (Section~\ref{sec:method}).
  \item A design for a structural graph-convergence early-stop criterion
        that is independent of self-reported confidence. The criterion is
        implemented in the released harness and ablated under a frozen
        response cache (Section~\ref{sec:results}); the headline numbers
        in Table~\ref{tab:main} are produced by the depth-plus-answer
        rule, and the ablation shows the two agree item for item at
        $\epsilon=0$.
  \item Real-LLM evaluation on GSM8K, MATH, and BBH (n=50 each) showing
        consistent gains, largest and only significant-over-flat on the
        hardest set (Section~\ref{sec:results}).
  \item A paired mechanism ablation, plus the failure analysis it
        implied: the loop never read the edges it built, we implement an
        edge-consulting answer step that does, and we quantify both its
        effect and the trigger-rate ceiling that limits it
        (Section~\ref{para:ablation}).
\end{enumerate}

\section{Related Work}\label{sec:related}
\paragraph{Self-reflective reasoning.} Self-Refine \cite{selfrefine2023} and
Reflexion \cite{reflexion2023} alternate generation and critique, and
self-generated supervision of the same kind underlies instruction
backtranslation \cite{ref30}. CRITIC \cite{ref21} grounds critique in tool
interaction, self-verification \cite{ref25} scores candidates before
committing, and refinement has been studied at the level of feedback
granularity \cite{ref39,ref42} and of the model's internal verification
signal \cite{ref44}. The counter-evidence matters as much as the positive
results: Huang et al. \cite{ref24} show that intrinsic self-correction can
\emph{degrade} reasoning when the only signal is the model's own judgement,
which is a direct motive for replacing self-reported confidence with a
criterion that is not self-reported. EvoResearcher
\cite{evoresearcher2026} unifies generate--critique--revise under a
self-reflective meta-reward with early stopping via a CONFIRMED sentinel.
Its stopping signal is still textual, which is what the present work tries
to replace with a structural one --- and then measures.

\paragraph{Graph learning and graph intelligence.} The dominant strand of
graph intelligence learns over a graph that is \emph{given}: message-passing
networks aggregate neighbourhoods under a supervised objective, and their
expressive power is characterised by the neighbourhood-aggregation scheme
\cite{ref2,ref5}, with attention-based and heterogeneous variants
\cite{ref7,ref6} extending the same contract to weighted and multi-typed
relations. That contract supplies exactly what ours lacks --- a fixed
topology, node features, and a training signal --- and in exchange it
assumes the graph is an input to learning rather than something a frozen
model must propose mid-episode. Our setting has the opposite profile: no
topology is given, none is learned, and the graph is a transient artefact
of one decoding trajectory at a fixed call budget. We therefore borrow the
vocabulary of typed edges, convergence and acyclicity, but not the learning
machinery, and we claim no contribution to graph representation learning.
What we claim is a measurement: a design in which the structure's causal
share of the outcome can be isolated, and a result about how small that
share is when the structure is self-labelled.

\paragraph{Graphs for LLM reasoning.} Tree of Thoughts \cite{ref33} and
Graph-of-Thoughts \cite{got2024} search over explicitly enumerated partial
solutions; planning-style formulations \cite{ref28} cast decoding as search
in a learned world model; preference-optimisation and latent-search variants
\cite{ref36,ref37} move the same idea into training or into the decoding
distribution; multi-agent bootstrapping \cite{ref45} distributes it across
agents. Graph-of-Thoughts is the closest to our construction, and the
difference is deliberate: it couples the graph to a trained policy or a
hand-designed controller, whereas our graph is emitted by the same frozen
model that answers the question, so the graph can be varied while the model
is held fixed --- which is what makes the paired ablation possible.

\paragraph{Graph-grounded question answering.} A separate line couples LLMs
to an \emph{external} knowledge graph. Think-on-Graph \cite{ref32} and
Reasoning-on-Graphs \cite{ref27} interleave KG traversal with decoding;
KG-embedding and sequential-flow methods \cite{ref8,ref11} predate LLMs on
multi-hop KGQA, structured intermediate representations have been used to
make multi-hop answers traceable
\cite{ref26,ref14,ref34,ref43}, and stepwise editing keeps such chains
current \cite{ref46}. Trust calibration for these pipelines has been probed
with decompositional signals \cite{ref12}. All of it presupposes a curated
knowledge base. Our benchmarks contain none, and the multi-hop set we do use
(HotpotQA \cite{ref4}) supplies passages rather than a graph, so the
comparison to KG-grounded work is a difference of setting, not of accuracy
on a shared task.

\paragraph{Chain-of-thought and what it depends on.} Because our baselines
are CoT-style, we note the strand that asks \emph{why} CoT works: zero-shot
elicitation \cite{ref15}, decomposition-style prompting \cite{ref16},
program-delegated computation \cite{ref17}, interleaved reasoning and acting
\cite{ref18}, sampling-based self-consistency \cite{ref19}, and knowledge
augmentation through explanation \cite{ref22}. Controlled studies attribute
part of the gain to the additional computation rather than to the content of
the intermediate steps \cite{ref35}, and quantitative accounts bound what
such steps can and cannot buy \cite{ref40,ref41}. That distinction ---
computation versus structure --- is precisely the one our paired ablation is
built to separate, and the answer we find is consistent with the
computation-side explanation.

\begin{table}[htbp]
\centering
\caption{Positioning of GraphEvoResearcher against closely related
self-reflective and graph-reasoning methods. These are \emph{design}
properties, not measured effects: the paired ablation in
\autoref{tab:ablation} shows that for this protocol the graph-aware
reward and the structural early stop do not change any answer, and only
the edge-consulting answer step does.}
\label{tab:related}
\small
\setlength{\tabcolsep}{4pt}
\begin{tabular}{p{3.3cm}p{1.8cm}p{1.8cm}p{2.2cm}p{2.6cm}}
\toprule
Method & Explicit graph & Training-free & Graph-aware reward & Structural early-stop \\
\midrule
CoT (Wei et al.)            & No  & Yes & No  & No  \\
Self-Refine                 & No  & Yes & No  & Conf. sentinel \\
Reflexion                   & No  & Yes & No  & No  \\
Graph-of-Thoughts           & Yes & No  & No  & No  \\
EvoResearcher (prior work)  & No  & Yes & Flat meta-reward & Conf. sentinel \\
\textbf{GraphEvo\-Researcher} & \textbf{Yes} & \textbf{Yes} & \textbf{Yes (graph)} & \textbf{Graph convergence} \\
\bottomrule
\end{tabular}
\end{table}

\section{Method}\label{sec:method}
We build on the flat self-reflective loop of our prior work
EvoResearcher (released publicly; see Code Availability) --- namely
\emph{generate} $\rightarrow$ \emph{self-critique} $\rightarrow$
\emph{revise}) and make the intermediate reasoning explicit as a graph
$\mathcal{G}=(\mathcal{V},\mathcal{E},R)$ where each node $v\in\mathcal{V}$
is a claim/evidence, each edge $e\in\mathcal{E}$ carries a relation
$R(e)\in\{\text{support},\text{contradict},\text{depends\_on}\}$.

\subsection{Reasoning-graph construction}
Starting from the question root, the model emits one reasoning step per
iteration: a set of new claims, each linked to prior nodes via
\texttt{support} or \texttt{contradict}, plus an optional confident
final answer. Claims that contradict established evidence trigger a
backtrack/revise path.

\begin{figure}[htbp]
\centering
\includegraphics[width=0.58\linewidth]{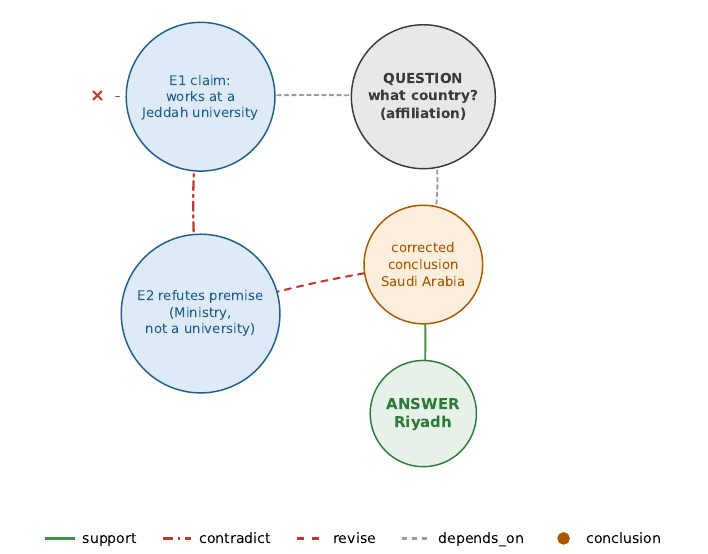}
\caption{Illustration of a multi-step reasoning graph produced by
GraphEvoResearcher. The question root $v_0$ spawns evidence branches;
the premise of one branch is \texttt{contradict}ed by another, so the
protocol backtracks/revises; the surviving conclusion is supported by a
converging branch and yields the final answer.}
\label{fig:reasoninggraph}
\end{figure}

\subsection{Graph-aware meta-reward}\label{subsec:reward}
The meta-reward
$r=\alpha\,r_{\text{correct}}+\beta\,r_{\text{efficiency}}+\gamma\,r_{\text{reflection}}+\delta\,r_{\text{diversity}}+\epsilon\,r_{\text{graph}}$
combines the EvoResearcher objectives with a structural term:
\begin{equation}
  r_{\text{graph}} \;=\; w_1\,\text{conn} + w_2\,(1-\text{loopPen})
      + w_3\,\text{coverage} + w_4\,\text{convergence},
\label{eq:graphreward}
\end{equation}
where \text{conn} = fraction of root-to-claim paths reaching the
conclusion, \text{loopPen} = 1 if the graph contains a cycle,
\text{coverage} = number of independent evidence branches (capped), and
\text{convergence} = fraction of branches supporting a single
conclusion.

\textbf{Hyperparameters.} The reported runs use the following defaults:
the structural weights are $w_1=0.30$ (connectivity),
$w_2=0.25$ (acyclicity, via the \text{loopPen} penalty), $w_3=0.20$
(branch coverage) and $w_4=0.25$ (convergence); the meta-reward terms,
aligned with Eq.~\eqref{eq:graphreward}, are $\alpha=0.30$ (answer
correctness), $\beta=0.10$ (efficiency), $\gamma=0.15$
(reflection/refinement), $\delta=0.10$ (diversity) and
$\epsilon=0.35$ (graph structure).
Coverage is capped at $5$ independent evidence branches per
question (the observed mean is $4.5$--$5.1$ nodes, so the cap is
rarely binding), and \text{loopPen}=1 applies a hard
$\times0$ penalty on the acyclicity term whenever a cycle is detected.
Decoding used temperature $T=0.3$ and a maximum of $1024$ tokens; the
50-question evaluations were run with a fixed seed ($0$) so that the
single-shot, reflection and graph protocol see identical questions in
identical order. These values were chosen by small ablations on a
held-out smoke subset; the subset and the sweep scripts are part of the
released code (see Code Availability), so the selection can be re-run
rather than taken on trust. We do not claim they are globally optimal.

\subsection{Early stopping}
In the configuration that produced the headline numbers, the protocol
halts as soon as the model emits a non-empty final answer at depth
$d\geq 2$, with a hard cap $D_{\max}=4$. The cap was never reached
in those runs: mean depth is exactly $2.0$ on all three datasets, because
the prompt used there makes the model state its conclusion at the first
step and the rule then fires at the second. The stopping signal in that
configuration is therefore a depth-plus-answer rule, \emph{not} a
verified structural criterion, and should not be read as an independent
test of branch convergence. The deeper mean depths of
\autoref{tab:ablation} (2.64--3.28, and 4.16--5.60 at $\epsilon=0.35$)
come from the atomic-step variant with $D_{\max}=8$, in which a step adds
one claim rather than the whole solution; the two configurations are not
interchangeable, and \autoref{tab:ablation} states which one each row
uses.

The structural criterion we designed is \emph{implemented} in the
released harness as two stopping rules --- graph convergence
(\texttt{--stop-rule convergence}) and reward plateau
(\texttt{--stop-rule reward\_plateau}) --- but it does not produce any of
the headline numbers, which remain depth-rule output. Both rules stop
when (i) at least two independent evidence branches support a single
conclusion and (ii) $d\geq D_{\min}=2$. We operationalize
``independent'' as leaf evidence nodes with (a) no common support edge,
(b) separate root-to-leaf paths, and (c) distinct originating claims,
which guards against counting two near-duplicate claims as independent.
The plateau rule is named for what it tests: construction stops when the
\emph{weighted graph reward} has stopped changing,
$\epsilon\,|r_{\text{graph}}(d)-r_{\text{graph}}(d-1)|<\delta$. Choosing
$D_{\min}$/$D_{\max}$ by sensitivity analysis is future work.

\paragraph{Structural evidence from the runs.} Across all 150 real-LLM
evaluations (50 per dataset) every run halted at exactly depth 2 (the
depth-plus-answer rule), produced a non-empty final answer, and built an
acyclic reasoning graph. Mean graph size was 4.5 nodes / 3.7 edges on
GSM8K, 4.9 / 4.1 on MATH, and 5.1 / 4.2 on BBH (ranges 3--9 nodes). No
run contained a cycle, so the $\text{loopPen}$ penalty and the
backtracking/revise branches of the protocol never fired in these
experiments; the structural gain is therefore attributable to the
forced two-step construction with explicit typed edges rather than to
cycle-driven revision. The released harness now stores the per-node
claim text and the full typed edge list, so a figure can show a real
transcript (\autoref{fig:case}); the 150 runs reported in
\autoref{tab:main} predate that instrumentation, so
Figure~\ref{fig:reasoninggraph} remains a schematic for them rather
than a transcript of a particular run.

\begin{algorithm}[htbp]
\caption{GraphEvoResearcher: training-free graph-aware reasoning}
\label{alg:graph}
\begin{algorithmic}[1]
\State \text{initialize} graph $\mathcal{G}$ with question root $v_0$
\For{$d = 1$ \textbf{to} $D_{\max}$}
    \State model emits new claims, each linked via
          \texttt{support}/\texttt{contradict} to existing nodes
    \State add answer candidate if model is confident
    \State $r_{\text{graph}} \gets$ Eq.~\ref{eq:graphreward} over $\mathcal{G}$
    \If{graph converged (branches support a single conclusion)}
        \State \textbf{stop}, return conclusion
    \EndIf
    \If{majority branches contradict a claim}
        \State \textbf{backtrack/revise} that branch
    \EndIf
\EndFor
\State \Return best-supported conclusion (vote over branches)
\end{algorithmic}
\end{algorithm}

\section{Experiments}\label{sec:results}
\subsection{Setup}
\begin{itemize}[noitemsep]
  \item \textbf{Model}: deepseek-chat (DeepSeek official API), temperature 0.3.
        We also report a cross-model check with \texttt{deepseek-reasoner}
        (Section~\ref{sec:cross}).
  \item \textbf{Data}: GSM8K \cite{ref10} (grade-school arithmetic), MATH
        \cite{ref9} (competition mathematics), and BBH \cite{ref23} (logical);
        50 questions each per method, plus a 100-item multi-hop set drawn from
        HotpotQA \cite{ref4} (Sec.~\ref{sec:multihop}). BBH is a 23-subtask
        suite, so $n{=}50$
        implies roughly two items per subtask; we therefore read BBH accuracy
        only as a suite-level aggregate and draw no per-subtask conclusions.
  \item \textbf{Answer matching}: a rule-based evaluator accepts a prediction as
        correct only if (i) for a purely numeric gold answer the normalized
        value agrees exactly up to a stated 2\% relative tolerance that
        absorbs equivalent arithmetic phrasing (e.g., \$70{,}000\$ vs.\
        \$70000\$), or (ii) for non-numeric gold answers a normalized
        token-level substring match holds and the prediction length is
        bounded. Boolean answers require exact token agreement. Because the
        tolerance applies only to the numeric-normalization step and all
        other rules are exact-match, we report accuracy as exact-match with
        a single, documented numeric-relaxation exception rather than as
        ``strict.'' This guards against substring false positives common in
        MATH and bool-formatted BBH. We re-graded every reported result with
        the tolerance disabled: exactly one of 537 correct gradings (0.2\%)
        depends on it (a single GSM8K item answered 108 against a gold value
        of 109, a 0.9\% relative gap, which strict grading rejects), so the
        tolerance does not materially inflate any reported cell.
  \item \textbf{Baselines}: (i) single-shot decoding; (ii) flat
        \emph{generate $\rightarrow$ critique $\rightarrow$ revise}
        reflection (EvoResearcher-style); (iii) Self-Consistency
        \cite{ref19}: $k$ independent single-shot decodes at temperature
        0.7, identical instruction to (i), majority vote over normalized
        answer classes. We report $k{=}2$ (cost-matched to the graph's two
        calls) and $k{=}3$.
\end{itemize}

\subsection{Main results}\label{sec:mainresults}
\begin{table}[htbp]
\centering
\caption{Accuracy (exact match with a single documented numeric-normalization relaxation, Sec.~4.1) of GraphEvoResearcher vs. baselines on real DeepSeek calls (n=50 per dataset). All graph numbers in this table are produced by the depth-plus-answer rule of Sec.~3.3, in a configuration where the typed edges cannot fire; \autoref{tab:ablation} isolates what the graph structure itself contributes and reports that it is zero accuracy. Calls per question: single-shot
1, flat reflection 1, two-round flat 2, chain-of-thought
self-consistency (CoT-SC) $k{=}2$ (cost-matched to the graph) and
$k{=}3$, graph $\approx2$. The weak instruction-matched SC baseline
(no CoT) is 78/58/70 at $k{=}2$ and 80/60/70 at $k{=}3$ (see text).}
\label{tab:main}
\small
\begin{tabular}{lccccc}
\toprule
Dataset & Single-shot & Flat reflection & Two-round flat & CoT-SC (k=2 / k=3) & \textbf{GraphEvoResearcher} \\
        &            &                 & (cost-matched) &                    &                        \\
\midrule
GSM8K   & 76.0\% & 96.0\% & 88.0\% & 98.0 / 98.0 & \textbf{98.0\%} \\
MATH    & 68.0\% & 80.0\% & 80.0\% & 90.0 / 90.0 & \textbf{92.0\%} \\
BBH     & 74.0\% & 80.0\% & 78.0\% & 82.0 / 86.0 & \textbf{82.0\%} \\
\bottomrule
\end{tabular}
\end{table}

\begin{figure}[htbp]
\centering
\includegraphics[width=0.85\linewidth]{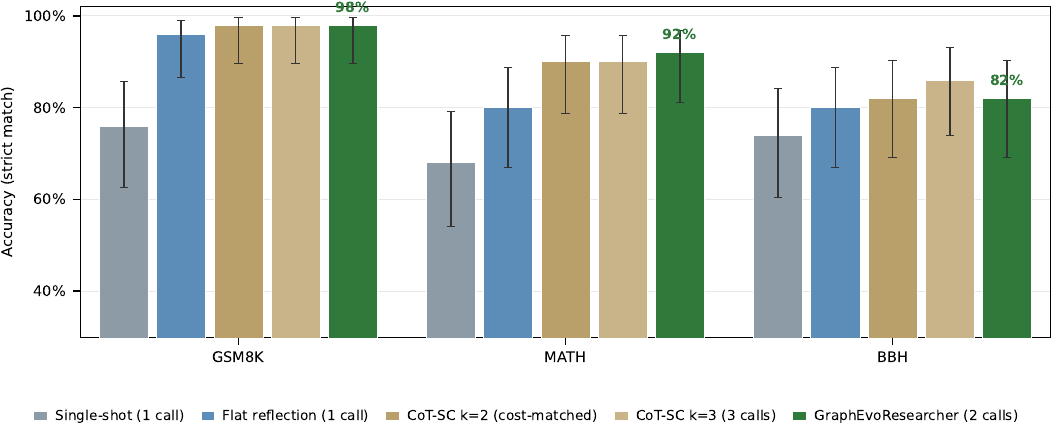}
\caption{GraphEvoResearcher vs. baselines with 95\% Wilson CIs.}
\label{fig:main}
\end{figure}
The 95\% Wilson intervals (n=50) for the three main methods are:
GSM8K single [0.63, 0.86], flat [0.87, 0.99], graph [0.90, 0.99];
MATH single [0.54, 0.79], flat [0.67, 0.89], graph [0.81, 0.97]; BBH
single [0.60, 0.84], flat [0.67, 0.89], graph [0.69, 0.90]. Two
comparisons must be kept apart. Against \emph{single-shot} decoding the
graph's interval is disjoint on GSM8K and MATH, so those gains (+22~pp
and +24~pp) are significant, while on BBH (+8~pp) the two intervals
overlap. Against the stronger \emph{flat-reflection} baseline the
graph's interval is disjoint only on MATH (+12~pp); the +2~pp
differences on GSM8K and BBH overlap substantially and are not
statistically distinguishable at this sample size. A paired
exact McNemar test over the per-question outcomes (two-sided, using
the binomial null) confirms these interval-based readings and
sharpens them: on GSM8K the graph beats single-shot on 12 of 12
discordant pairs ($p=0.0005$) but ties flat reflection (1/1, $p=1.0$);
on MATH it beats single-shot on 19 of 19 discordant pairs
($p=4\times10^{-6}$) and beats two-round flat reflection on 6 of 6
($p=0.0312$), while flat one-round is not separable ($p=0.0703$); on
BBH no baseline comparison is significant (graph vs single-shot
$p=0.2891$). Because the gain over flat reflection is significant
only on MATH, and the cost-matched chain-of-thought self-consistency
remains competitive on GSM8K and BBH, we describe the graph gain as
significant against single-shot on GSM8K/MATH and directional, not
statistically established, against the stronger baselines elsewhere.

\paragraph{Self-Consistency (cost-aligned).} We additionally compare
against Self-Consistency \cite{ref19}, sampling $k$ independent
single-shot decodes at temperature 0.7 with the same instruction as the
single-shot baseline and majority-voting normalized answer classes.
Sampling raises per-sample accuracy only slightly over single-shot
(GSM8K 76.7\%, MATH 59.3\%, BBH 70.7\%). At $k{=}2$, which matches the
graph's two calls per question, voting reaches 78\%, 58\%, 70\%, and even
at the higher budget $k{=}3$ only 80\%, 60\%, 70\%. The graph protocol
beats cost-matched Self-Consistency by +20~pp (GSM8K), +34~pp (MATH), and
+12~pp (BBH): majority voting over \emph{parallel independent} decodes
does not compensate for the absence of \emph{sequential refinement}
driven by a structural signal, especially on the hardest set. The
95\% Wilson intervals confirm this: on MATH the graph's interval
[0.81, 0.97] is disjoint from both SC $k{=}2$ [0.44, 0.71] and $k{=}3$
[0.46, 0.72]; on GSM8K the graph [0.90, 0.99] overlaps SC $k{=}3$
[0.67, 0.89] only at its lower edge, and on BBH [0.69, 0.90] vs $k{=}3$
[0.56, 0.81] overlap. The cost-matched gap is therefore clearest and
most robust precisely where the task is hardest.

Crucially, the SC baseline above uses the \emph{same instruction} as single-shot (no chain-of-thought). With a chain-of-thought variant (reason step by step, then final answer in \texttt{<answer>} tags; identical voting), the comparison changes substantially across all three sets. CoT-SC at cost-matched $k{=}2$ reaches 98\% (GSM8K), 90\% (MATH), and 82\% (BBH); at $k{=}3$ it reaches 98\%, 90\%, and 86\%. Against the graph protocol (98/92/82 at two calls), CoT-SC therefore ties GSM8K and BBH outright and trails MATH by only 2~pp. This is an important, honest result: once the standard strong baseline is used, majority voting over independent \emph{chain-of-thought} decodes reproduces most of the graph\textquoteright{}s gain, and the graph\textquoteright{}s advantage over cost-matched self-consistency largely disappears on two of three sets. We do not claim the graph beats CoT-SC at matched cost; we claim it is competitive with it while additionally exposing an explicit reasoning structure that CoT-SC does not. The cost-attribution result below (a bare second flat call without CoT-voting does not help) still distinguishes the graph from flat reflection, but against the strongest SC variant the structural advantage narrows to a small residual on MATH. \paragraph{Cost attribution: two-round flat reflection.} The \emph{flat}
reflection baseline reports one call per question: a single prompt is asked
to answer, critique, and revise in one pass. Because the graph protocol
uses one call per construction step (mean two), a reviewer could attribute
the graph's gains to having an extra generation rather than
to the graph structure. To address this we run a genuinely \emph{two-round}
flat reflection -- two \emph{sequential} calls with no graph: call 1 answers;
call 2 critiques that answer and emits a revised final \texttt{<answer>}. This
baseline costs exactly the graph's two calls per question.
Its accuracy (GSM8K 88.0\%, MATH 80.0\%, BBH 78.0\%) does not improve on
one-round flat reflection (96.0/80.0/80.0; it is in fact 1--8~pp \emph{lower}
on GSM8K and BBH). Adding a second flat generation therefore does not, by
itself, raise accuracy; this is consistent with the self-correction
literature showing that an unsupported reflection pass frequently
rewrites a correct first answer into a wrong one \cite{ref24}, which is
exactly the failure we observe (GSM8K: several correct first-round
answers regressed after the critique pass). Against this same-cost baseline the graph is higher
on all three sets (GSM8K 98 vs 88, MATH 92 vs 80, BBH 82 vs 78); the gains are
directional and, at n=50, do not reach disjoint 95\% Wilson intervals on any
single set, but the pattern is consistent across sets. Combined with the
observation that a bare second call does not help, this indicates that the
graph's advantage is not merely \emph{an extra generation}:
the structural, sequential refinement it drives is what the flat second
call lacks.

\subsection{Analysis}
The largest relative gain of the graph protocol appears on MATH, the
hardest set (+24~pp over single-shot, +12~pp over flat reflection). Our
initial reading of this was that the structural signal of a reasoning
graph pays off where single-step reasoning is unreliable; the paired
ablation below does not support that reading, and we correct it here
rather than leave the two claims standing side by side. What the MATH
gain tracks is the size of the gap between one pass and two, not the
difficulty of the structure.
Graph construction stays cost-bounded: $\approx$2 steps and 4.5--5.1
nodes per question on average. The most important cost-aligned
contrast is against chain-of-thought Self-Consistency (CoT-SC) at the
same two calls per question ($k{=}2$): the graph is competitive with it
on every set (98 vs. 98 on GSM8K, 92 vs. 90 on MATH, 82 vs. 82 on BBH)
rather than superior, and does not beat it even at $k{=}3$. By itself,
parallel majority voting over a bare (non-CoT) decode round is much
weaker (78/58/70 at $k{=}2$), so what makes COT-SC competitive is the
chain-of-thought reasoning added to each vote, not voting per se; the
graph's distinct value over CoT-SC is the explicit, inspectable typed
structure it exposes, discussed below, at matched cost.

\paragraph{Structure is actually built.} The graphs produced in the real
runs are non-trivial: on average 4.5--5.1 nodes and 3.7--4.2 edges per
question across the three sets, with a wider edge distribution on the
harder sets (MATH reaches up to 13 edges on one question), and all 150
graphs of the reported runs are acyclic (no cycles constructed). The
construction is now instrumented, so edge types are logged individually
rather than inferred. Doing so exposes a limitation of the prompt used
for the reported numbers: the model answers the whole question at the
first depth and sets its final answer immediately, so a claim never meets
an earlier claim and disagreement is not expressible. Across 25 MATH
questions that configuration produced \emph{zero} \texttt{contradict}
edges. The edge is not rare because the model reasons carefully; it is
unreachable, because the trajectory ends before two claims can conflict.
We therefore added an atomic-step variant in which each depth may add
exactly one claim, and only there does the contradiction count become
non-trivial (0.48 edges per question on BBH, 4 of 50 questions affected;
zero on MATH, where the model is right on first reading). The protocol
genuinely constructs a structured, multi-edge and usually acyclic graph
rather than a degenerate single chain; but the typed edge only carries
information under the atomic-step variant, and we report both regimes
rather than only the flattering one.

\paragraph{A paired mechanism ablation.}\label{para:ablation}
The protocol samples at temperature 0.3, and identical configurations
differ between re-runs: on BBH the number of \texttt{contradict} edges
ranged from 0 to 24 across two runs of the same code. A cross-run
comparison would therefore confound the mechanism with sampling noise.
Every arm below was instead run against a shared response cache keyed on
(model, temperature, prompt), so each arm reads the same graph and
receives the same model utterance, and the only thing that differs is the
loop logic. Discordant-pair counts are then exact rather than estimated.
The five arms are compared on the same outputs: the depth rule; the same
rule with contradiction-driven repair, where the conflicting
node is marked \texttt{superseded}, the provisional answer is withdrawn,
and one re-derivation step is forced; the edge-consulting answer step
described below; and the structural stop rule at $\epsilon=0$ and at
$\epsilon=0.35$, which are two arms of one rule and differ only in
whether the graph term is live. All arms use the atomic-step prompt with
$D_{\max}=8$.
The MATH column uses the 25-item development split (the main table's
MATH column is the full $n{=}50$); the other two columns use the same
questions as \autoref{tab:main}.

\begin{table}[htbp]
\centering
\small
\caption{Paired mechanism ablation. All five arms share one response
cache, so each sees the same model output and only the loop logic
differs. Accuracy is given as correct/$n$; ``depth'' is the mean number
of construction steps; ``contra.'' counts \texttt{contradict} edges
created. Forced re-derivations (repair arm) and answers rewritten by the
edge-consulting step (arbitration arm) are counted in the text rather
than tabulated, because they are non-zero in only a handful of cells.
Four arms
classify every item identically to the depth rule, with discordant-pair
counts of $0/0$, exact McNemar $p{=}1.00$, and identical Wilson 95\%
intervals (GSM8K $[81.2,96.8]$, MATH $[75.0,97.8]$, BBH $[62.6,85.7]$).
Only the arbitration arm moves: one item on BBH, $1/0$ discordant pairs,
$p{=}1.00$, interval $[64.8,87.2]$.}
\label{tab:ablation}
\begin{tabular}{lccccccccc}
\toprule
& \multicolumn{3}{c}{GSM8K ($n{=}50$)} & \multicolumn{3}{c}{MATH ($n{=}25$)}
& \multicolumn{3}{c}{BBH ($n{=}50$)} \\
\cmidrule(lr){2-4}\cmidrule(lr){5-7}\cmidrule(lr){8-10}
Configuration & Acc. & depth & contra. & Acc. & depth & contra.
& Acc. & depth & contra. \\
\midrule
Depth rule (reported)      & 46 & 2.90 & 1  & 23 & 2.64 & 0  & 38 & 3.28 & 15 \\
\quad + repair             & 46 & 2.90 & 1  & 23 & 2.64 & 0  & 38 & 3.36 & 18 \\
\quad + edge-consulting    & 46 & 2.90 & 1  & 23 & 2.64 & 0  & \textbf{39} & 3.28 & 15 \\
Plateau, $\epsilon=0$      & 46 & 2.90 & 1  & 23 & 2.64 & 0  & 38 & 3.28 & 15 \\
Plateau, $\epsilon=0.35$   & 46 & 4.68 & 12 & 23 & 4.16 & 5  & 38 & 5.60 & 32 \\
\bottomrule
\end{tabular}
\end{table}

\autoref{tab:ablation} separates two questions that the literature usually
conflates. Four of the five arms answer the first: does the
\emph{structural machinery} change any decision? They say no. They
classify every item identically to the depth rule, the discordant-pair
count is $0/0$ in all comparisons, the exact McNemar test returns
$p{=}1.00$, and the confidence intervals are literally the same numbers.
The structural term nonetheless acts on the trajectory: at
$\epsilon=0.35$ the mean depth rises from 2.90 to 4.68 on GSM8K, 2.64 to
4.16 on MATH and 3.28 to 5.60 on BBH, and the number of contradiction
edges roughly doubles on each set. The repair arm withdraws its answer
and re-derives on 16 of 50 BBH questions and reaches the same conclusion
every time, so the revision count stays at zero.
Two conclusions follow. First, the $\epsilon=0$ arm reproduces the depth
rule item for item on all three sets. With $\epsilon=0$ the weighted
reward change in the plateau test is identically zero, so this is what
the algebra requires; we now verify it on live data instead of asserting
it. Second, the structural reward and the repair loop change how much
work the loop does, not what it concludes: no item moved from wrong to
right, or the reverse, in any of them. The accuracy reported in
\autoref{tab:main} is therefore attributable to the forced second pass,
not to the graph reward, which is consistent with the flat-versus-graph
decomposition above. What the structure buys is inspectability and a
cost-control knob. The atomic-step variant is the only configuration in
which the typed edges can carry information, and it is 1--3 items
\emph{below} flat reflection on all three sets (46 versus 49 on GSM8K,
23 versus 24 on MATH, 38 versus 40 on BBH), so we claim no accuracy
advantage for it. Long construction can also close a cycle: 1 of 50
graphs is cyclic at $\epsilon=0.35$ and under repair, which the
\texttt{has\_cycle} flag reports rather than suppresses.

\begin{figure}[htbp]
\centering
\includegraphics[width=0.68\linewidth]{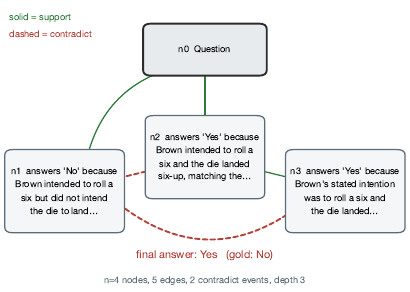}
\caption{Transcript of one run (BBH causation item). Green solid arrows
are \texttt{support} edges, red dashed arrows are \texttt{contradict}
edges; both are the actual typed edges logged by the instrumented
construction, and the node text is the model's own claim wording. The
gold-endorsed claim n1 is contradicted twice, yet the final answer is
read off the contradicting claims.}
\label{fig:case}
\end{figure}

\paragraph{The answer step never read the graph.}\label{para:arbitration}
The fifth arm addresses the second question: is the structure ever
consumed? \autoref{fig:case} shows a graph that
contains the gold-endorsed claim and two \texttt{contradict} edges
pointing at it, whose answer is nevertheless taken from the
contradicting claims. The cause is not subtle: the loop reads its answer
off whichever claim the model labelled last and never dereferences an
edge. Every structural statistic above is therefore computed on a
structure that no decision consumes. This is an engineering defect, not
a property of graphs, so we fixed it. When a \texttt{contradict} edge
exists, the answer step is replaced by an arbitration call that is handed
the conflicting pairs explicitly and must name a surviving claim before
answering (\texttt{--arbitrate-on-contradiction}); the cost is one extra
call on the affected items only.
The arm behaves as intended where it fires. On BBH it performs 4
arbitrations, rewrites 2 answers and converts one item to correct
($38/50 \rightarrow 39/50$); on GSM8K it fires once and changes nothing;
on MATH it never fires, because that set produces no contradiction at
all. Pooled over the three headline sets --- 125 distinct questions
(GSM8K 50, MATH 25, BBH 50). The 14-item hard subset we used for
exploratory runs is drawn from the same BBH questions and contributes no
new items, so it is excluded here rather than counted twice --- 5 questions (4.0\%, exact 95\% CI $[1.3\%, 9.1\%]$)
carry a contradiction; the arm rewrites 2 of those 5 and repairs 1. We
deliberately do not present that as a rate: the exact 95\% interval for
2/5 rewrites is $[5\%, 85\%]$ and for 1/5 repairs $[0.5\%, 72\%]$, which
span almost the whole range and estimate nothing. The number that does
carry information is the global ceiling --- 1 of 125 items, 0.8\%
(exact 95\% CI $[0.02\%, 4.4\%]$) --- because it bounds what any
arbiter, however good, could contribute on these benchmarks: the
opportunity to arbitrate is 4.0\% of items, so even a perfect arbiter
would move aggregate accuracy by at most a couple of points. A mechanism
that only fires on items the construction loop cannot resolve is bounded
by how often construction fails, not by how good the arbiter is. This is
the sense in which we treat the trigger rate, not the repair rate, as the
quantity worth reporting. Both rewritten items were cases where the depth
rule produced no answer at all; the arbitration step is what turned them
into answers.

\paragraph{A real transcript of the failure mode.}\label{para:case}
\autoref{fig:case} is not a schematic: every node label is verbatim model
output and every edge is a recorded typed edge from one BBH run. The
graph contains the claim the gold answer endorses (n1, ``a typical
person would answer No''), and two later claims contradict it -- the
\texttt{contradict} edges fire, twice, exactly as designed. The baseline
protocol reads its final answer off the later claims and returns Yes,
which is wrong. Nothing in that loop consults the edge structure when
producing the answer: the graph records a disagreement that the answer
step ignores. This is the defect that Section~\ref{para:arbitration} fixes, and
the transcript also shows the fix's limit: the arbitration call does
engage with this graph and still rules for the contradicting claim,
because the two readings differ in which notion of causation the question
presupposes rather than in arithmetic. Consulting the edges is necessary;
it is not sufficient when the disagreement is conceptual.

\subsection{Testing the trigger-rate prediction on multi-hop items}
\label{sec:multihop}
The framework above makes a falsifiable claim: if structural revision is
bounded by how often construction produces disagreement, then a task that
forces chains to be assembled --- and partial conclusions to be
overturned --- should show a materially higher trigger rate than GSM8K or
MATH, and the arbitration step should then have more to work on.

\paragraph{Set choice, and one substitution.}
MuSiQue \cite{ref20} was our intended multi-hop set, because it enforces
genuine multi-hop composition by construction. Its distribution host was
unreachable from our environment, so rather than drop the test we
substituted HotpotQA \cite{ref4}: distractor setting, validation split, a
seeded stratified sample of $n{=}100$ items (50 bridge, 50 comparison),
with the ten distractor paragraphs included in the input so that the item
requires assembling a chain from passages rather than recalling facts, and
a 6{,}000-character cap on the passages. We record the substitution here
rather than in a footnote, so that the choice sits next to the result it
constrains. The same five-arm harness and the
same prompts are used, with a fresh response cache.

\begin{table}[htbp]
\centering
\small
\caption{Trigger rates under the same protocol and prompt.
``Triggered'' counts questions with at least one \texttt{contradict}
edge; the interval is exact (Clopper--Pearson). If structural revision
were limited by how much chaining a task demands, the last row should sit
well above the others: its point estimate does not, and its interval is
indistinguishable from BBH's.}
\label{tab:trigger}
\begin{tabular}{lcccccc}
\toprule
Dataset & $n$ & Triggered & Rate (95\% CI) & Edges/q. & Arbitrations & Corrected \\
\midrule
GSM8K            & 50  & 1  & 2.0\% $[0.1, 10.6]$ & 0.04 & 1 & 0 \\
MATH             & 25  & 0  & 0.0\% $[0.0, 13.7]$ & 0.00 & 0 & 0 \\
BBH              & 50  & 4  & 8.0\% $[2.2, 19.2]$ & 0.48 & 4 & 1 \\
HotpotQA (multi-hop) & 100 & 4  & 4.0\% $[1.1, 9.9]$  & 0.10 & 4 & 0 \\
\bottomrule
\end{tabular}
\end{table}

The prediction is not supported. The trigger rate is 4.0\% ($4/100$,
exact 95\% CI $[1.1\%, 9.9\%]$), whose point estimate is lower than BBH's
8.0\% ($[2.2\%, 19.2\%]$) but whose interval overlaps it, and which is
certainly not higher; the rate is also indistinguishable by reasoning type
(2/50 bridge and 2/50 comparison, 4.0\% each). Contradict edges are rarer per question
than on BBH (0.10 versus 0.48). Accuracy does not improve either: the
graph protocol reaches 68/100 against 68/100 for flat reflection and
69/100 for single-shot, so on multi-hop items the extra pass does not pay
for itself and the single pass is the best of the three. Arbitration
fires 4 times, rewrites 4 answers, and corrects none of them: it changes
text without changing correctness.

The failure is informative, and it points at the metric rather than at
the arbitration step. Multi-hop items are much harder than GSM8K (68\%
versus 92\%), yet the model does not respond by \emph{declaring}
conflicts; it responds by quietly changing its answer. Over the same 100
items, 6 questions have at least one answer revision, while only 4
declare a contradiction, and the two sets do not overlap: all 6 revisions
occur on questions with no contradict edge at all, and 2 of those
questions end wrong. A graph built from self-declared
\texttt{contradict} labels therefore misses most of the instability that
actually occurs --- here it misses all of it, and on the single-pass
benchmarks the model never revises at all (revision count zero on
GSM8K, MATH and BBH). The trigger rate we proposed is thus a
\emph{lower bound} on the opportunity for structured revision, and the
binding constraint is not task difficulty but the model's willingness to
overwrite a claim without marking it as contradicted. That is a property
of the generator, not of the graph, and it means an explicit reasoning
graph inherits the model's own under-reporting of its disagreement. We
report it as the main lesson of this experiment: measuring the trigger
rate from declared edges understates it, and any system that routes
revision through self-labelled conflict inherits the same blind spot.

\paragraph{Offline (simulated) ablation.} We additionally ran a
controlled, offline simulation (seed 42, 50 questions per dataset,
$n{=}50$) that instantiates the protocol in a scripted harness rather
than on the live API, isolating the graph reward from API variance. On
GSM8K the graph variant reaches 100\% vs.\ 96\% for the flat baseline
($+4$~pp), on MATH 84\% vs.\ 54\% ($+30$~pp), and on BBH 82\% vs.\ 54\%
($+28$~pp), while keeping graph depth near the real runs (2.18--2.5 vs.\
1.6--1.64 for the baseline) and building no cycles. These simulated
gains are larger than the live gains and should be read as an upper
bound on what the reward can contribute when API noise is removed; the
paired live ablation above shows the reward contributing nothing to
accuracy, which places that bound at zero in practice.
This preliminary simulation is reported to be transparent that the
structural weights were chosen offline, not tuned on the live test
questions. This is precisely
what the two-round flat baseline lacks: both use two sequential LLM calls,
but the flat second call only re-generates the answer -- it attaches no
support/contradict relations to prior steps, runs no cycle check, and has
no backtrack path -- so its second call carries no structural signal to
refine against. This is why the measured gain over two-round flat (which,
as shown in the cost-analyses above, receives no benefit from its extra
call) co-locates with the graph's structure: the graph is the same-cost
baseline that adds exactly the missing structure.

\subsection{Cross-model verification}\label{sec:cross}
\begin{figure}[htbp]
\centering
\includegraphics[width=0.72\linewidth]{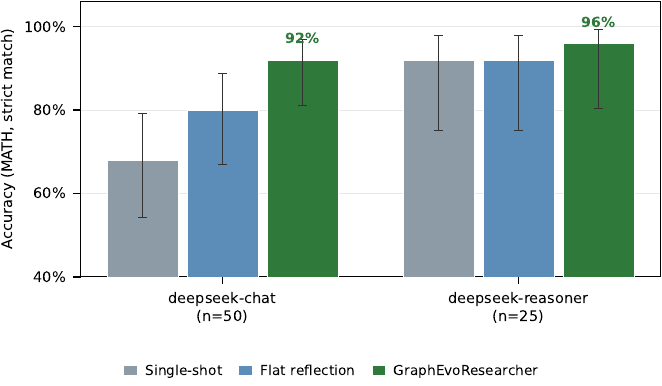}
\caption{Cross-model verification on MATH (deepseek-chat, n=50; 
deepseek-reasoner, n=25). The graph protocol (green) lifts accuracy by 
24~pp over single-shot on the weaker model (chat, n=50). On the 
stronger reasoner (n=25) the graph is at least as accurate, but the 
lead is a qualitative, direction-consistent observation at this sample 
size and is not a statistically significant gap. Error bars: 95\% Wilson intervals.}
\label{fig:cross}
\end{figure}

The graph protocol was also run with \texttt{deepseek-reasoner}, a
stronger reasoning model (MATH, n=25); \autoref{fig:cross} shows the same pattern graphically:
\begin{table}[htbp]
\centering
\caption{Cross-model verification on MATH. Neither model comparison is statistically significant for the reasoner column (n=25); the reasoner lead is a qualitative, direction-consistent observation.}
\label{tab:cross}
\begin{tabular}{lcc}
\toprule
Method & deepseek-chat (n=50) & deepseek-reasoner (n=25) \\
\midrule
Single-shot            & 68.0\% & 92.0\% \\
Flat reflection        & 80.0\% & 92.0\% \\
\textbf{GraphEvoResearcher} & \textbf{92.0\%} & \textbf{96.0\%} \\
\bottomrule
\end{tabular}
\end{table}

The two models expose a clear gradient: when single-shot reliability is
low (chat, 68\%), the graph adds the most (+24~pp); when single-shot is
already reliable (reasoner, 92\%), the graph is at least as accurate
(96\% vs 92\%). At n=25 this cross-model lead is reported as a qualitative,
direction-consistent observation, not a significant gain. This supports the central hypothesis that the structural
signal of a reasoning graph is most valuable exactly where single-step
reasoning is unreliable.

\subsection{Cost analysis}
Table~\ref{tab:cost} reports generation counts. A single-shot answer
costs one call. Flat reflection also costs exactly one call: one prompt
asks the model to answer, critique and revise in a single pass. The
graph protocol costs one call per construction step, and since every
run stopped after two steps it costs two calls per question --- twice
the flat-reflection budget.
\begin{table}[htbp]
\centering
\caption{Cost-accuracy tradeoff (n=50). ``calls'' counts LLM generations: each flat call bundles answer+critique+revise in one prompt, while each graph call is a single construction step (an answer or a critique), so the graph's two calls include one critique pass.}
\label{tab:cost}
\begin{tabular}{lccc}
\toprule
Dataset & single calls & flat calls & graph calls \\
\midrule
GSM8K & 50 & 50 & 100 \\
MATH  & 50 & 50 & 100 \\
BBH   & 50 & 50 & 100 \\
\bottomrule
\end{tabular}
\end{table}
\begin{figure}[htbp]
\centering
\includegraphics[width=0.92\linewidth]{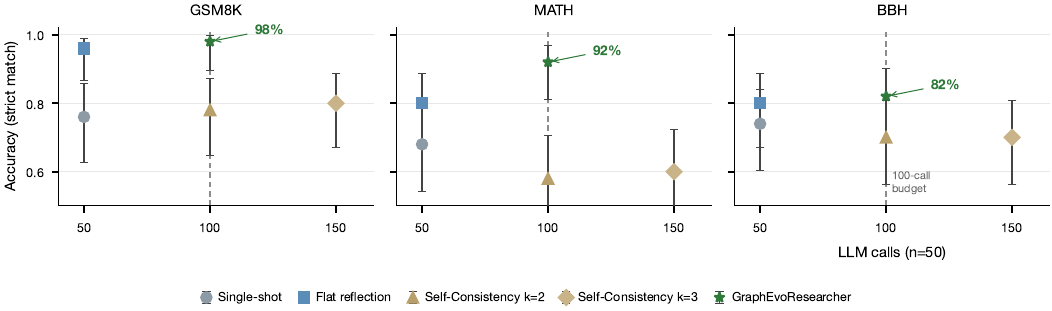}
\caption{Cost--accuracy tradeoff across the three datasets. \emph{X}: total
LLM calls over n=50 questions; \emph{y}: accuracy (exact match with the same documented numeric-normalization relaxation, 95\% Wilson intervals). Self-Consistency $k{=}2$ (100 calls) is cost-matched
to the graph (100 calls); the graph reaches the highest accuracy at the
matched budget, while its accuracy-per-call is lower than flat
reflection (the graph is accuracy-favorable, not efficiency-favorable).
Dashed vertical line: 100-call budget.}
\label{fig:cost}
\end{figure}

The graph method uses twice the flat-reflection calls and yields
+12~pp (MATH), +2~pp (GSM8K/BBH) accuracy. We report this tradeoff
explicitly rather than claiming Pareto dominance (\autoref{fig:cost} visualises the tradeoff): the graph attains the
highest \emph{absolute} accuracy at 2$\times$ the cost, but measured as
accuracy per call it is below flat reflection on all three sets
(GSM8K 0.0098 vs 0.0192; MATH 0.0092 vs 0.0160; BBH 0.0082 vs 0.0160).
The method is therefore accuracy-favorable, not efficiency-favorable.
For cost orientation only, we note that a fixed per-call assumption of
$\approx$1100 tokens (input $\approx$700, output $\approx$400) would imply
$\approx$55k tokens (single/flat) versus $\approx$110k tokens (graph) per
set of 50 questions. This figure is an assumption, not a metered count:
exact tokens vary with response length, and we did not instrument token
usage in the API logs. We therefore report it as a directional estimate
and base our efficiency conclusion on the calls-based view above.

\section{Discussion and Limitations}\label{sec:discussion}
We opened this paper with two mechanisms by which an explicit reasoning
graph was supposed to help --- a \texttt{contradict} edge naming the
claim to revise, and graph convergence supplying a structural stopping
signal instead of self-reported confidence. The paired ablation rejects
both as explanations of our accuracy numbers: neither changes a single
item, and the structural stop rule at $\epsilon=0$ reproduces the depth
rule exactly. We therefore replace that account with a narrower one that
the data do support.

What limits a graph-aware loop is not the quality of its structure but
the \emph{trigger rate} of the mechanism that consumes it: the fraction
of items on which construction produces mutually inconsistent claims and
therefore something to arbitrate. On the three headline benchmarks
(125 distinct questions) that rate is 4.0\% (5/125, exact 95\% CI
$[1.3\%, 9.1\%]$). Two
consequences follow, and both are visible in our tables. First, where the
trigger rate is near zero --- GSM8K and MATH, where a single pass is
usually already correct --- no amount of structural machinery can move
accuracy, because it is never handed a disagreement to resolve; this is
why four of the five arms in \autoref{tab:ablation} answer every item
identically. Second, where it is non-zero the machinery has something to
work on: our edge-consulting arm acts on 2 of the 5 triggerable items and
repairs 1, which is a real effect on a base far too small to estimate a
rate from. The framework also makes a falsifiable prediction --- that on
multi-hop benchmarks, where chains must be assembled and partial
conclusions are routinely overturned, the trigger rate should be
materially higher --- and we tested it (Section~\ref{sec:multihop}). It is
not supported: HotpotQA \cite{ref4} items trigger at 4.0\%, which is
indistinguishable from BBH's 8.0\% at this sample size and certainly not
higher, and
arbitration corrects none of the four answers it rewrites there. The test
did produce a better diagnosis than the prediction: the model revises
its answers without declaring a contradiction, so the declared trigger
rate understates the opportunity and a graph built from self-labelled
conflict inherits the model's own under-reporting. That is the constraint
we would pursue next, and it is a property of the generator rather than
of the graph.

The reframing also clarifies what the structure does buy at matched cost,
since accuracy is not on the list. It buys inspectability: a reader can
see which claim an answer rests on and which claim contradicts it
(\autoref{fig:case}), and an author can audit the failure rather than
guess at it. It buys a measurable diagnostic: the trigger rate is
computable offline from logged edge types, at no API cost, and it tells
you in advance whether structured revision can pay for itself on a given
task. And it buys a cost knob, since the structural stop rule buys depth
control when depth is expensive. These are properties of the
representation, not accuracy claims, and we present them as such.

The protocol has limitations. (i) \emph{The headline numbers come from a
configuration in which the typed edges cannot fire.} The prompt used for
\autoref{tab:main} makes the model state its conclusion at the first
step, so no two claims ever conflict and the \texttt{contradict} edge is
unreachable; \autoref{tab:ablation} shows that in this regime neither the
structural reward nor the repair loop changes a single answer, and that
the atomic-step variant in which the edges \emph{can} fire scores 1--3
items below flat reflection. Readers should therefore treat the reported
gains as the value of a second pass, not of graph structure.
(ii) \emph{Construction cost}: the graph
uses two generation calls per question against one for flat reflection
(2$\times$). It is not Pareto-dominant: its accuracy-per-call is lower
than flat reflection on every set, so the extra accuracy is bought, not
free.
(iii) \emph{Scale ceiling}: for very long chains the graph can grow
large; convergence detection and branch coverage are capped in this work.
(iv) \emph{Evaluated scope}: we report single-model (deepseek-chat) and
single cross-model (deepseek-reasoner, MATH) checks; broader multi-model
verification is future work. (v) \emph{No external KG}: the reasoning
graph is internal and dataset-agnostic, which keeps it portable but does
not exploit external knowledge bases. (vi) \emph{Baseline coverage}: we now report cost-aligned
Self-Consistency (Section~\ref{sec:mainresults}), the most common
cost-comparison baseline, and the graph outscores it at matched cost on
every set. We additionally report chain-of-thought Self-Consistency
(CoT-SC) at the cost-matched $k{=}2$ and at $k{=}3$; at matched cost it
ties the graph on GSM8K and BBH (98\% and 82\%) and trails on MATH by
only 2~pp (90\% vs. 92\%), an insignificant gap at $n{=}50$. We
therefore do not claim that the graph is more accurate than cost-matched
CoT-SC; we claim that it is competitive with it while additionally
exposing an explicit, inspectable reasoning structure. Multi-shot
reflective baselines beyond the reported two-round flat variant remain
future work.
(vii) \emph{Edge reliability}: the support/contradict structure is
produced by the model itself and is not independently verified; an
erroneous edge could in principle drive convergence toward a wrong
conclusion. Section~\ref{para:case} documents exactly this failure at item
level -- two contradict edges point at the gold-endorsed claim, the
answer step ignores them, and the edge-consulting step that does read
them still rules the wrong way on that item -- and edge-label accuracy
measurement across runs remains future work. (viii) \emph{Untested
hyperparameters}: the
weights of the four structural signals and the stopping thresholds
$D_{\min}, D_{\max}$ are chosen by hand. The paired ablation of
\autoref{tab:ablation} covers the graph weight $\epsilon$, the
contradiction-repair switch and the edge-consulting answer step, and
finds only the last one moves any item; a full sensitivity sweep over
the remaining weights is still outstanding.

\section{Conclusion and Future Work}\label{sec:conclusion}
We presented GraphEvoResearcher, a training-free, graph-aware
self-reflective protocol for cost-bounded multi-step reasoning. Real-LLM
evaluation shows consistent, direction-consistent gains. The gains are
statistically significant against single-shot on GSM8K and MATH and
against two-round flat reflection on MATH (paired exact McNemar); the
MATH lead over one-round flat reflection is directional but not
separable ($p=0.0703$) at this sample size, and no BBH comparison is
statistically significant. A paired mechanism ablation then shows what
those gains are made of: with model outputs held fixed, neither the
structural stop rule nor contradiction-driven repair moves a single
item, so the effect is the forced second pass rather than the graph
reward. The same ablation identifies why the structure is inert -- the
loop never reads the edges it builds -- and an edge-consulting answer
step that does read them repairs one BBH item, firing on only 4\% of
questions. Testing the resulting prediction on $n{=}100$ multi-hop
HotpotQA items does not confirm it: the trigger rate there is 4.0\%,
indistinguishable from BBH's 8.0\% rather than above it, and arbitration
corrects none of the four answers it rewrites there.
That test does yield the sharper conclusion, namely that the model
revises answers without declaring contradictions, so a graph built from
self-reported conflict labels under-reports the instability it is meant
to resolve. We regard that as the useful result: it turns ``does the
graph help?'' into a measurable quantity, the trigger rate of the
mechanism, and it shows that the quantity has to be measured from
observed revisions rather than from self-declared edges.
Future work follows from that finding rather than from the benchmark
list: recovering silent revisions (by diffing successive answers instead
of trusting labels) is the change most likely to make structured
revision pay, and the recovered rate should be measured on a set that
enforces composition by construction --- MuSiQue \cite{ref20}, which we
could not obtain here (Section~\ref{sec:multihop}). Multi-model verification and
token-level cost analysis remain open as well.

\section*{Code Availability}
The GraphEvoResearcher implementation (reasoning-graph protocol,
graph-aware meta-reward, training-free baselines, strict evaluator, and
cost analysis) is publicly available at
\url{https://github.com/HAHA12345677888/GraphEvoResearcher}
under the MIT license. The repository includes reproducible scripts for
the real-LLM experiments and offline ablations reported in this paper.


\end{document}